\documentclass[a4paper]{article}

\usepackage{INTERSPEECH2022}
\usepackage{array}
\usepackage[hidelinks]{hyperref}
\usepackage{breakurl}
        \hypersetup{
           breaklinks=true,   
           pdfusetitle=true,  
        }

\title{Toward Fairness in Speech Recognition: \\
Discovery and mitigation of performance disparities}
\name{
Pranav Dheram, Murugesan Ramakrishnan, Anirudh Raju, I-Fan Chen,  Brian King,   \\ 
Katherine Powell, Melissa Saboowala, Karan Shetty, Andreas Stolcke
}
\address{
  Amazon Alexa AI, USA
  }
\email{pddheram@amazon.com, ranirudh@amazon.com, ifanchen@amazon.com}

\begin{document}

\maketitle

\begin{abstract}
As for other forms of AI, speech recognition has recently been examined with respect to performance disparities across different user cohorts. One approach to achieve fairness in speech recognition is to (1) identify speaker cohorts that suffer from subpar performance and (2) apply fairness mitigation measures targeting the cohorts discovered. In this paper, we report on initial findings with both discovery and mitigation of performance disparities using data from a product-scale AI assistant speech recognition system. We compare cohort discovery based on geographic and demographic information to a more scalable method that groups speakers without human labels, using speaker embedding technology.  For fairness mitigation, we find that oversampling of underrepresented cohorts, as well as modeling speaker cohort membership by additional input variables, reduces the gap between top- and bottom-performing cohorts, without deteriorating overall recognition accuracy.
\end{abstract}

\section{Introduction}

Customers and society expect AI-based products to perform equally well for a wide range of users across gender, race, and ethnic backgrounds. While failures in this regard have received wide attention mainly in consideration of computer vision applications \cite{pmlr-v81-buolamwini18a,Mehrabi21,raji2020saving,de2019does} and natural language processing \cite{hutchinson2020social,pruksachatkun2021does,liang2020towards}, we should be similarly concerned about fairness in speech-based AI systems, in the specific sense that we want equally high accuracy for users regardless of speaker-based attributes, be they geographical or demographic background, or speaker and voice characteristics that may not even be readily interpretable but are predictive of speech recognition performance. 
In this paper we specifically look at performance fairness for automatic speech recognition (ASR) systems.

Our approach to ASR fairness consists of two stages. First, we aim to identify cohorts of speakers that suffer from subpar ASR accuracy; we call this stage {\em (fairness) cohort discovery}.
Second, we deploy methods to reduce such performance disparities, e.g., by changing the composition of the training data or applying better modeling techniques;
we call this second stage {\em (fairness) mitigation}.
In this study we report initial results for both cohort discovery and mitigation on a production-scale, de-identified dataset representative of ASR used for a deployed AI assistant.
For cohort discovery, we contrast two approaches. The first is based on human demographic labels and geographic proxy information (ZIP codes), while the second uses machine learning based on speaker embeddings that quantify speaker (voice) similarity. We find that the ML-driven approach is not only more scalable but also identifies more significant discrepancies for larger cohorts of underserved speakers.

For fairness mitigation, we investigate two well-known methods from the literature.
We find that oversampling training data and modeling of cohort membership in the ASR model each reduce the relative word error rate (WER) gap between the top- and bottom-performing cohorts derived from ZIP codes, by 17.9\% and 31.6\% respectively, as measured for an end-to-end neural ASR system. These reductions in performance disparity are achieved without significant reductions in top-cohort accuracy.

Research in fairness for speech recognition  is still in its nascent stage, but is of great importance to society given the increasing pervasiveness of ASR technology. Prior studies have shown ASR performance disparities based on gender, age, race and ethnic backgrounds \cite{Koenecke20, heldreth2021don, feng2021quantifying, liu2021towards, riviere2021asr4real, bajorek}.
We hope to advance the work in this area with three contributions:
(1) An assessment of fairness in speech recognition at scale; (2) a novel approach for automated discovery of underperforming cohorts; and (3) to the best of our knowledge, a first report on fairness mitigation for production-scale ASR systems.

\section{Human and Machine performance disparities}
While prior work \cite{garnerin2019gender} has discussed representation bias in training data, we also consider transcription bias as a potential source of disparity. Specifically, we hypothesize that underperforming cohorts are not just difficult for machines (machine accuracy) but also for human annotators (human accuracy). Consequently, this could affect the quality of labels, in turn affecting ASR performance. 
In this section, we first discuss how human and machine disparities are quantified across cohorts and then explore two approaches to define top and bottom cohorts.
Here we refer to ``bottom cohort'' as an identifiable subset of speakers with average performance metric worse than the average; ``top cohort'' is its complement.
In Section~\ref{sec:results}, we present both human and machine disparities across these cohorts discovered using the two approaches on de-identified data. 
 
\subsection{Machine performance disparities}
We consider two metrics of ASR model performance: (1) average model confidence score (higher is better) \cite{swarup2019improving} and (2) WER (lower is better).  WER computation, while giving a more accurate metric of ASR correctness, requires reference transcripts. The model confidence score, which is the estimated probability of an output sequence being correct, removes the dependence on reference labels, facilitating assessment on larger datasets. 

\subsection{Human performance disparities}
Standard WER compares the ASR hypothesis to a reference, usually generated by human
transcribers. References are not error-free, therefore the measured WER is an aggregate of errors both in the ASR and in the reference. For our study, we generate human transcriptions for a de-identified internal dataset through a multipass process similar to \cite{glenn-etal-2010-transcription}. 
First, three annotators transcribe an utterance independently, referred to as ``blind passes''. The transcription system then compares these transcriptions. If two or more of the blind passes match for an utterance, that becomes the final transcription version. If none of the blind passes match, the utterance will be transcribed by a fourth and final person. In this adjudication pass, the transcriber listens to the audio and can choose a transcript from one one of the blind passes, or create a fourth unique transcription. The three options for blind pass agreement are thus 1-1-1 (all three disagree), 2-1 (two agree), and 3-0 (all three agree).

We hypothesized that utterances that are challenging for an ASR system to recognize are often also challenging for humans, in agreement with findings that human and machine speech recognition error rates are highly correlated at a speaker level \cite{stolcke17_interspeech}. In order to test this, we analyzed inter-annotator agreement rates for the utterances in our top- and bottom-cohort evaluation datasets.
If a dataset had more agreement (more 3-0 and 2-1), then it was easier for the transcribers to transcribe. If it had less agreement (more 1-1-1), then it was more challenging.

\subsection{Geodemographic cohort discovery}
\label{sec:geo-demographic}
One approach to define cohorts is via geographic and public demographic information.
Since we do not have access to individual speakers' demographic information, we use
geolocation in a ZIP code approximated using device location as a means to categorize on the basis of demographics. We identify ZIP codes with a high percentage of a demographic category as a proxy attribute. (The demographic information comes from US census data.)
For any such category, we consider all speakers in one of the ZIP codes where that category comprises 75\% or more of the population as a potential bottom cohort; all other ZIP codes form the top cohort.  We then select for top/bottom cohorts that yield high performance disparities.

\vspace{-1mm}
\subsection{Automatic cohort discovery}
    \label{sec:automatic-cohort}

While relying on geolocation and associated demographics yields some measure of  interpretability it also poses challenges for accuracy and scalability.  Specifically, geodemographic cohorts selection as used here
\begin{itemize}
    \item is limited to census categories, which may not be relevant to speaker differences impacting ASR performance;
    \item typically yields very skewed top/bottom splits, limiting the share of speakers to which mitigation can be applied;
    \item does not scale to locales for which the geodemographic data infrastructure (census data, location-mapping) is not available;
    \item has limited resolution and therefore accuracy, since all speakers within a ZIP code are lumped together. Also, the available census data may not be representative of device users within a ZIP code.
\end{itemize}
To avoid these limitation we investigate automated cohort discovery methods that rely on measurable data at the speaker level, with a special focus on data that is directly relevant to speech recognition. In such a data-centric approach, we propose to use machine learning models that predict ASR performance from observable speech features. To identify these speech features, we predict ASR performance based on speaker embeddings. We first extract speaker embeddings from the wake word segment (the initial portion) of an utterance using a pretrained speaker embedding model. The speaker embedding extractor is an LSTM network trained using generalized end-to-end loss \cite{wan2018generalized}. Embeddings thus extracted from utterances are further aggregated by speaker.

We then train an unsupervised K-means model on the aggregated speaker embeddings to identify clusters of similar-sounding speakers. After experimenting with different numbers of means (clusters) K, we identified 50 clusters as sufficient to represent speech characteristics relevant to ASR. The ASR model’s average confidence score is used to rank the speaker clusters by ASR difficulty. The 10\% of clusters with the lowest confidences are defined as the bottom cohort, while the remaining clusters form the top cohort.
We expect these clusters of speaker embeddings to capture voice characteristics that are relevant to ASR difficulty.

\section{Mitigation Methods}

Here, we discuss methods that are meant to reduce the performance gaps between cohorts, found using either one of the methods described below, or otherwise.

\subsection{Semi-supervised learning (SSL): oversampling bottom cohort data}

Production-scale speech recognition systems are typically trained on a mix of human-transcribed and semi-supervised data. Our chosen SSL method utilizes machine-generated transcripts from a stronger speech recognition teacher model. We hypothesize that one of the root causes of performance degradation on the bottom cohort data is that it is underrepresented in the training data. For example, statistics from our ASR training data show that less than 1\% of the training data is from the bottom cohort ZIP codes. This mitigation method focuses on a data selection scheme that oversamples training data from the bottom cohort ZIP codes and utilizes it in semi-supervised training. A benefit of this method is that it does not require any additional human annotation for either data selection
or transcription, as this would be infeasible at the scale required for model training. This approach could also be applied to cohorts from automatic cohort discovery. 

\subsection{Cohort embeddings}
It is observed in the dialect speech recognition community that having a simple one-hot accent embedding as an additional input to the acoustic model input layer provides significant accuracy improvement for the accented speech; the performance is comparable to adaptation approaches with much larger complexity \cite{Grace18}. The one-hot embedding serves as an adapting bias in the first layer of the model for the given accent. We adopt this idea to our situation by creating a 2-dimensional one-hot cohort embedding representing top and bottom cohorts as an additional input
to the ASR model, concatenated with acoustic features. The idea is that the cohort embedding will allow the model to adapt to linguistic differences between top and bottom cohorts, thereby
reducing disparity.

\begin{table}[t]
\centering
\caption{Hybrid RNN-HMM ASR model performance disparity (\textbf{confidence scores relative to overall confidence score on the dataset}) in production-scale data on the basis of geodemographic characteristics.}
\vspace{-1em}
\label{tab:machine-disparity-confidence-manual}
\begin{tabular}{|c|c|c|c|}
\hline
Geolocation-based & \#ZIPs & \#Hrs (K) & ASR Conf. \\
Cohorts                     & & & Relative \\
\hline
Overall            & 41696  & 4513 & Baseline \\ 
Bottom          & 431  & 48 & -11.7\% \\ 
\hline
\end{tabular}
\end{table}

\begin{table}[t]
\centering
\caption{WER Gap for two cohort discovery approaches (\textbf{different datasets were used for both}); \% of bottom cohort data in \textit{overall population for each approach} is also shown. ASR Model used: Hybrid RNN-HMM.}
\vspace{-1em}
\label{tab:WER-Gap}
\begin{tabular}{|c|c|c|}
\hline
Cohort Discovery               & WER-gap (\%) & Bottom cohort \\
    &   & share (\%) \\
\hline
Geodemographic            & 41.7 & 0.8 \\ 
Automatic            & 65.0 & 10.0 \\ \hline
\end{tabular}
\end{table}

\begin{table}[t]
\centering
\caption{Inter-annotator agreement rates of the three blind passes for geodemographic differences based cohorts.}
\vspace{-1em}
\label{tab:transcription}
\begin{tabular}{|c|c|c|c|c|c|}
\hline
Cohorts               & 1-1-1 & 2-1 & 3-0 & PDR & Rel. PDR \\
 & (All 3 & (Two & (All 3 & & increase \\
 & disagree) & agree) & agree) & & in Bottom \\
\hline
Top            & 16.9\%  & 32.2\% & 51.0\% & 27.6\% & - \\ 
Bottom            & 26.2\%  & 33.4\% & 40.5\% & 37.3\% & 35.1\% \\
\hline
\end{tabular}
\end{table}

\begin{table}[t]
\centering
\caption{Inter-annotator agreement rates of the three blind passes for the automatically detected cohorts.}
\vspace{-1em}
\label{tab:human-disparity-automated}
\begin{tabular}{|c|c|c|c|c|c|}
\hline
Cohorts               & 1-1-1 & 2-1 & 3-0 & PDR & Rel. PDR\\
              &(All 3& (Two & (All 3 & & increase \\
               &disagree) & agree) & agree) & & in Bottom \\
\hline
Top            & 11.8\%  & 24.6\% & 63.6\% & 20.0\% & - \\ 
Bottom            & 16.7\%  & 30.0\% & 53.5\% & 26.7\% & 33.5\% \\ \hline
\end{tabular}
\end{table}

\section{Experimental Setup}

\textbf{Evaluation datasets}. We created evaluation datasets from de-identified far-field US English data spoken to a commercial voice-enabled artificial intelligence assistant. These are used to measure both accuracy and cross-cohort discrepancies for baseline and new models. We chose a single census attribute that gave the highest gap in ASR confidence after mapping ZIP codes with at least 75\% majority with that attribute to the bottom cohort, as described in Section~\ref{sec:geo-demographic}. Finally, we applied filters selecting utterances ranging between Jan '21 and May '21, removing the 3.9\% of lowest-confidence, typically noise, utterances from both cohorts, as well as removing utterances consisting of only a
wake-word. The bottom cohort comprises 2hrs of data and the top cohort dataset has 31hrs. We evaluated several production-scale ASR models on these datasets to establish a baseline WER and WER gap. We also compared the distributions of several high-level attributes (including domain, intent and signal-to-noise ratios), so as to rule them out as potential sources of discrepancies.

\textbf{Training datasets}. A semi-supervised learning (SSL) training dataset for the bottom cohort was prepared by selecting utterances based on geolocation and obtaining machine transcriptions from a teacher model. We used a bidirectional-LSTM teacher model which had better individual WERs as well as WER-gap compared to our baseline model. 

\textbf{Baseline model}: As a baseline, we use a recurrent neural network transducer \cite{Graves13,Graves12} (RNN-T) model. The 39M-parameter model comprises 4 encoder layers and 2 decoder layers (each LSTM layer with 512 hidden units) and was trained on 100,000s of hours of voice-assistant data. The acoustic features used by the RNN-T are 64 log-mel filterbank energies computed over a 25\,ms window with 10\,ms shift, stacked with two frames to the left and downsampled to a 30\,ms frame rate.

\textbf{Metrics of interest}: 
1) \textit{WER-gap}: measures  the relative degradation of WER on the bottom cohort, compared to the top cohort, using the same ASR model:
\[\textit{WER-gap} = \frac{\textit{WER}_\textit{bottom} - \textit{WER}_\textit{top}}{\textit{WER}_\textit{top}}
\]
2) \textit{Word error rate reduction (WERR)}: We report our candidate model's reduction in Word Error Rate for a given cohort with respect to baseline's performance on the same cohort
\[\textit{Bottom WERR} = \frac{\textit{WER\_Baseline}_\textit{bottom} - \textit{WER\_Candidate}_\textit{bottom}}{\textit{WER\_Baseline}_\textit{bottom}}
\]
3) \textit{Pairwise disagreement rate (PDR):\footnote{Other inter-annotator agreement metrics such as Cohen's kappa \cite{cohen1960coefficient} normalize for agreement by chance. Since chance agreement probability is a constant in all our comparisons we report the raw rates of (dis)agreement here.}}
captures the likelihood of disagreement between two randomly selected transcribers in transcribing an utterance from the dataset of interest. We expect a larger disagreement rate on more difficult dataset. Relative increase in PDR reflects how much more likely transcribers are to disagree on the bottom cohort in comparison to the top cohort. In the below formula $\textit{Percentage}_\textit{1-1-1}$ refers to the percentage of times all three transcribers disagreed and $\textit{Percentage}_\textit{2-1}$ reflects the percentage of times two transcribers agreed but one disagreed.
\[
\textit{PDR} = \textit{Percentage}_\textit{1-1-1} \cdot \frac{3}{3}+\textit{Percentage}_\textit{2-1} \cdot \frac{2}{3}
\]

\begin{table*}[t]
\centering
\caption{Relative WER-gap between top and bottom cohorts, showing impact of SSL and cohort embeddings. Word error rate reduction (WERR) compared with baseline experiment E0 is reported, with positive values indicating improvements.}
\label{tab:results}
\begin{tabular}{|c|c|c|c|c|c|}
\hline
Exp.No & Model               & Relative WER-Gap & Bottom WERR & Top WERR & \% Bottom cohort samples\\
    &           & (\%) & (\%) & (\%) & in training\\
\hline
E0 & Baseline            & 56.3 & 0 & 0 & 0.8 \\
E1 & E0 + SSL           & 46.2 & 5.0 & -1.6 & 9.0\\ 
E2 & E0 + Cohort embedding           & 38.5 & 10.0 & -1.6 & 0.8 \\ 
E3 & E0 + Cohort embedding + SSL           & 40.0 & 9.0 & -1.6 & 9.0 \\
\hline
\end{tabular}
\end{table*}

\section{Results}
    \label{sec:results}

\subsection{Cohort Discovery and measuring disparities}

\subsubsection{Geodemographic cohort discovery}

\textbf{Machine performance disparities.}
On a dataset comprising millions of hours of utterances, we first identified cohorts and measured model performance, using ASR confidence scores to avoid the need for transcriptions. Of the several cohorts identified on the basis of geodemographic similarities, we selected one distinct low-ASR-accuracy cohort based on a demographic attribute, as described in Section~\ref{sec:geo-demographic}, yielding the statistics shown in Table~\ref{tab:machine-disparity-confidence-manual}.
The resulting bottom cohort includes 431 ZIP codes; all remaining ZIP codes make up the top cohort.
The bottom cohort has an average confidence score that is 11.7\% lower than the overall average. On further analysis, the distribution of domains and intents was not significantly different between top and bottom cohorts, thus ruling out different usage patterns as an explanation of the performance discrepancy.
Work described later on mitigation will focus on improving model performance for this bottom cohort.

We also divided a smaller dataset with transcriptions into top and bottom cohort partitions, using the same division of ZIP codes, and computed WERs.
The results, given in row 1 of Table~\ref{tab:WER-Gap}, show disparities in machine transcription accuracy that align with the discrepancy in confidence scores.

\noindent\textbf{Human performance disparity.}
As shown in Table~\ref{tab:transcription}, when we compare agreement rates between Top and Bottom, we see that Top had higher 3-0, similar 2-1, and lower 1-1-1 agreement versus Bottom. In fact, the 1-1-1 agreement rate is 55\% relative higher on Bottom. As the Bottom dataset has a higher WER than Top, we may surmise that cohorts that ASR finds challenging are also more challenging for humans to transcribe.
The underlying reasons could well be similar, too: just as bottom cohort data is insufficiently represented in ASR training data, transcribers may not be sufficiently familiar with speech from an underserved cohort, purely based on how frequently such speech is encountered.

\subsubsection{Automatic cohort discovery}

\noindent\textbf{Machine performance disparities.}
 In Table~\ref{tab:WER-Gap}, we evaluate the ASR model on the automatically discovered top and bottom cohorts based on speakers' voice similarity. We observe that WER on the bottom cohort is about 65\% worse than on the top cohort. We notice that WER-gap is larger between the cohorts identified with this approach, compared to cohorts selected based on geodemographic information. Furthermore, the bottom cohort identified by 
 our automatic method represents a much larger share of the overall population (10\% versus 0.8\%).
Although the underlying datasets are not identical, they both have a similar data distribution, leading us to conclude that the automatic approach is more effective at identifying ASR disparities in terms of both WER gap and bottom cohort share.

In future work, we intend to employ different cohort discovery approaches on a common dataset and quantify WER disparity using a unified metric, such as Gini coefficient \cite{sen1997economic}. At present we do not have an analysis of the speaker characteristics of the automatically selected cohort, other than in terms of ASR performance.  While it would be interesting to have such insights and a measure of interpretability, we note that one of the strengths of the automatic method is precisely that it does not require demographic or geographic labels.
Machine-learning driven cohort discovery is free to discover disparities that are not necessarily reflected in interpretable attributes.
 
\noindent\textbf{Human performance disparities.}
In Table~\ref{tab:human-disparity-automated}, we compare human annotator agreement on the top and bottom cohorts based on the automatic discovery pipeline. Similar to results in Table~\ref{tab:transcription}, we note that the bottom cohort has more inter-annotator disagreements, and is thus more difficult to transcribe not just for machines, but for humans as well.

\subsection{Mitigation}
We report results on mitigating disparities only among the geodemographic cohorts discovered above.

\subsubsection{Oversampling bottom cohort data improves model performance}
In Table~\ref{tab:results}, we compare a baseline RNN-T ASR model with an experiment that oversamples bottom cohort data in training. This yields a 5\% relative improvement in bottom cohort WER and reduces the WER gap between top and bottom-performing cohorts from 56.3\% to 46.2\%.

\subsubsection{Impact of cohort embedding}
Table~\ref{tab:results} shows the experimental results. Both SSL and cohort
embedding approaches alone reduce the WER gap from 56.3\% to 46.2\% and 38.5\% respectively. However, combining the two shows no further gain. The result is counterintuitive, in that we would expect that with more bottom cohort data, the cohort embedding model would better adapt to
bottom-cohort accents and yield additional gap reduction. We hypothesize that this is due to the imperfect correlation between ZIP codes and spoken language varieties. 

To verify this hypothesis, we trained a cohort classifier predicting cohort labels given input speech audio and used it to relabel the utterances in the bottom SSL training data. We found that, with
a 0.75 classifier threshold, only 40\% of the utterances are classified as bottom cohort based on their acoustic features. This suggests only a portion of speech from bottom-cohort ZIP codes exhibit associated speech characteristics, confusing the cohort embedding model.

\section{Conclusion}

We have examined two key aspects of ensuring fair ASR performance in real-world systems, using data from a production-scale recognition system typical for deployed AI voice assistants.
For identifying speaker cohorts suffering from subpar ASR accuracy, we compared a method that partitions speakers based on US geographic areas (ZIP codes) and associated majority demographic attributes obtained from census data.  While this method uses interpretable attributes and does identify underperforming cohorts, it is by definition imprecise and not even applicable to locales lacking geolocation or demographic data.
We find that an automatic cohort discovery method that groups speakers by voice similarity, without reference to human labels, is able to identify larger performance discrepancies, and larger underperforming cohorts.
For mitigating ASR performance discrepancies, we experimented with two methods: oversampling of the target cohort using semi-supervised training, and modeling of cohort membership using additional model inputs. Both methods, when applied to a
geodemographic split chosen for high initial performance disparity, are effective in reducing the relative WER gap between top and bottom-performing cohorts from 56\% to below 40\%.

Further directions for research include  interpretable automatic cohort discovery solutions by analyzing quantities such as pitch, speaking rate and speaking rhythm. Further, we hope to expand our fairness mitigation methods to loss functions (e.g., counterfactual loss \cite{kusner2017counterfactual}) and modeling (e.g., adaptation techniques), applying them to automatically identified cohorts. 

\clearpage

\bibliographystyle{IEEEtran}

\bibliography{mybib}

\end{document}